\documentclass[journal]{IEEEtran}
\pdfoutput=1
\usepackage{times}
\usepackage{epsfig}
\usepackage{graphicx}
\usepackage{amsmath}
\usepackage{amssymb}
\usepackage{array}
\usepackage{stfloats}
\usepackage{diagbox}
\usepackage{multirow}
\usepackage{subfig}
\usepackage{url}
\usepackage{color}

\newcolumntype{L}[1]{>{\raggedright\let\newline\\\arraybackslash\hspace{0pt}}m{#1}}
\newcolumntype{C}[1]{>{\centering\let\newline\\\arraybackslash\hspace{0pt}}m{#1}}
\newcolumntype{R}[1]{>{\raggedleft\let\newline\\\arraybackslash\hspace{0pt}}m{#1}}

\ifCLASSINFOpdf
\else
\fi

\begin{document}
%
% paper title
\title{Feature Calibration Network for Occluded Pedestrian Detection}
\author{Tianliang~Zhang,~\IEEEmembership{Student Member,~IEEE,} Qixiang~Ye,~\IEEEmembership{Senior~Member,~IEEE,} Baochang~Zhang,~\IEEEmembership{Member,~IEEE}, Jianzhuang~Liu,~\IEEEmembership{Senior~Member,~IEEE}, Xiaopeng~Zhang, Qi~Tian,~\IEEEmembership{Fellow,~IEEE}
\thanks{T. Zhang is with the School of Electronics, Electrical and Communication Engineering, University of Chinese Academy of Sciences, Huairou, Beijing, 101408, China. E-mail: (zhangtianliang17@mails.ucas.ac.cn). This work was done when he was a visiting student at Huawei Noah's Ark Lab.}
\thanks{Q. Ye is with the School of Electronics, Electrical and Communication Engineering, University of Chinese Academy of Sciences, Huairou, Beijing, 101408 China. E-mail: (qxye@ucas.ac.cn). Q. Ye is the corresponding author.}
\thanks{B. Zhang is with Institute of Deep Learning, Baidu Research, Beijing, Email: (bczhang@139.com).}
\thanks{J. Liu is with Huawei Noah's Ark Lab., Shenzhen, China. E-mail: liu.jianzhuang@huawei.com.}
\thanks{X. Zhang and Q. Tian are with Cloud \& AI, Huawei Technologies, Shenzhen, China. E-mail: (zhangxiaopeng12@huawei.com, tian.qi1@huawei.com).}
}

\markboth{IEEE Transactions On Intelligent Transportation Systems}%
{Shell \MakeLowercase{\textit{et al.}}: Bare Demo of IEEEtran.cls for IEEE Journals}

\maketitle

\begin{abstract}
Pedestrian detection in the wild remains a challenging problem especially for scenes containing serious occlusion. In this paper, we propose a novel feature learning method in the deep learning framework, referred to as Feature Calibration Network (FC-Net), to adaptively detect pedestrians under various occlusions. FC-Net is based on the observation that the visible parts of pedestrians are selective and decisive for detection, and is implemented as a self-paced feature learning framework with a self-activation (SA) module and a feature calibration (FC) module. In a new self-activated manner, FC-Net learns features which highlight the visible parts and suppress the occluded parts of pedestrians. The SA module estimates pedestrian activation maps by reusing classifier weights, without any additional parameter involved, therefore resulting in an extremely parsimony model to reinforce the semantics of features, while the FC module calibrates the convolutional features for adaptive pedestrian representation in both pixel-wise and region-based ways. Experiments on CityPersons and Caltech datasets demonstrate that FC-Net improves detection performance on occluded pedestrians up to $10\%$ while maintaining excellent performance on non-occluded instances.
\end{abstract}

\IEEEpeerreviewmaketitle

\section{Introduction}
\IEEEPARstart{P}{edestrian} detection is an important research topic in the computer vision area, driven by many real-world applications including autonomous driving \cite{DBLP:conf/iccv/ChenSKX15}, video surveillance \cite{DBLP:journals/pami/HaritaogluHD00}, and robotics \cite{YeSelf2017,DBLP:conf/cvpr/ZhangBOHS16,PersonSearch2018}. With the rise of deep learning, pedestrian detection has achieved unprecedented performance in simple scenes. However, the performance for detecting heavily occluded pedestrians in complex scenes remains far from being satisfactory~\cite{wang2017repulsion, DBLP:conf/wacv/DuELD17,DBLP:conf/eccv/ZhangLLH16, DBLP:conf/iccv/BrazilYL17,DBLP:conf/cvpr/ZhangBS17, DBLP:conf/cvpr/MaoXJC17}. For example,  when the occlusion rate is higher than 35\% (Caltech pedestrian dataset), state-of-the-art methods \cite{DBLP:conf/iccv/BrazilYL17} report miss rates larger than $50\%$ at 0.1 False Positive Per Image (FPPI). This seriously hinders the deployment of pedestrian detection in real-world scenarios.

\begin{figure}[tbp]
    \centering
    \includegraphics[width=1.0\linewidth]{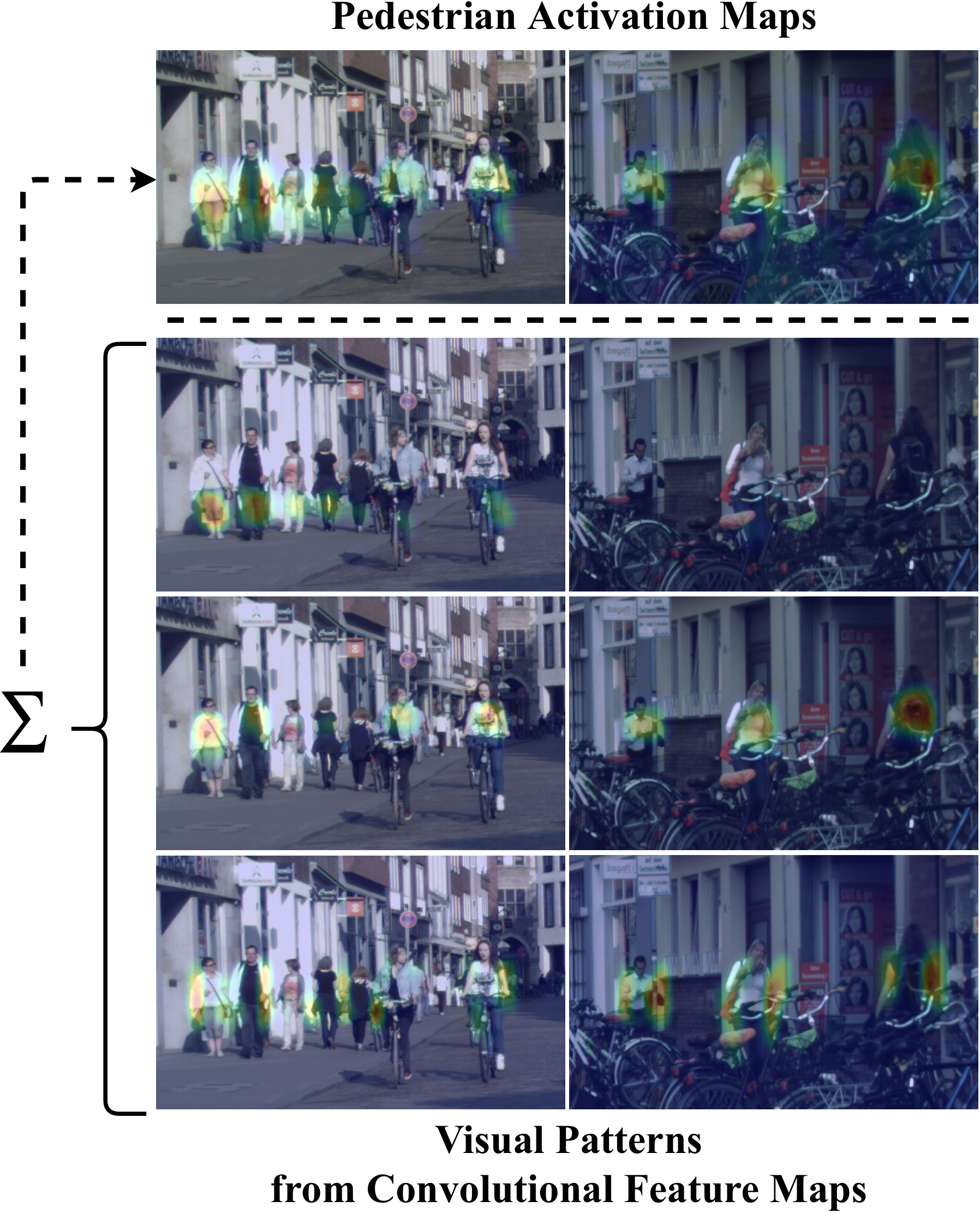}
    \caption{Pedestrian activation maps (upper) and visual patterns (lower). (Best viewed in color)}
     \label{fig_PAM}
\end{figure}

To address the occlusion issue, one commonly used method is the part-based model \cite{DBLP:conf/iccv/TianLWT15}, which leverages a divide-and-conquer strategy to handle visible and occluded parts.
However, such a method suffers deficiency in handling complex occlusions due to a limited number of parts and the fixed part partition strategy.
The other commonly used method is the attention model~\cite{zhang2018occluded}, which replaces ``hard" object parts with ``soft" attention regions by introducing feature enforcement and/or sampling modules \cite{SE-Net2018,DBLP:conf/iccv/BrazilYL17}.
Nevertheless, the attention model usually operates in parallel with the detector learning procedure, ignoring the class-specific semantic information produced by the detectors.
This could mix the attentive regions of negatives and positives and make the feature enforcement dubiously oriented.

In this paper, we propose self-activation (SA) and feature calibration (FC) modules, and target at adaptating the convolutional features to pedestrians of various occlusions. The SA module defines the corresponding relationship between pedestrians and convolutional feature channels, without any additional parameter involved. Such relationship is reflected by a classifier weight vector, which is constructed during the learning of the detection network. By multiplying such a weight vector with the feature maps in a channel-wise manner, the visual patterns across channels are collected and a pedestrian activation map is calculated, as shown in Fig.\ \ref{fig_PAM}. The activation map is further fed to the FC module to reinforce or suppress the convolutional features in both pixel-wise and region-based manners.

Integrating the SA and FC modules with a deep detection network leads to our feature calibration network (FC-Net). In each learning iteration, FC-Net updates the classifier weights, which are reused to calibrate the features, iteratively. The key idea of calibration is leveraging the pedestrian activation map as an indicator to reinforce the features in visible pedestrian parts while depressing the features in occluded pedestrian regions. With multiple iterations of feature calibration, FC-Net attentively learns discriminative features for pedestrian representation in a self-paced manner.

The contributions of this work include:

(1)	We propose a self-activation approach, and provide a simple yet effective way to estimate pedestrian activation maps by reusing the classier weights of the detection network.

(2)	We design a feature calibration module, and upgrade a deep detection network to the feature calibration network (FC-Net), which can highlight the visible parts and suppress the occluded parts of pedestrians.

(3)	We apply FC-Net on commonly used pedestrian detection benchmarks, and achieve state-of-the-art detection performance with slight computational cost overhead. We also validate the applicability of FC-Net to general object detection.

The remainder of the paper is organized as follows. In Section II, related work about pedestrian detection and occlusion handling is described. In Section III, the implementation details of the SA and FC modules are presented. In Section IV, the learning procedure of FC-Net for pedestrian detection is described. We show the experiments in Section V and conclude the paper in Section VI.

%-------------------------------------------------------------------------
\section{Related Work}

There is a long history of pedestrian detection research, and various feature representations have been proposed including Histogram of Gradients (HOG) \cite{DBLP:conf/cvpr/DalalT05, DBLP:journals/pami/FelzenszwalbGMR10}, Local Binary Patterns (LBP) \cite{DBLP:journals/pami/OjalaPM02}, Integral Channel Feature (ICF) \cite{DBLP:conf/bmvc/DollarTPB09}, \cite{DBLP:conf/icassp/KeZWYJ15}, and informed Haar-like features \cite{InforHaar2014, TowardsHuman18,DBLP:journals/pami/DollarABP14}. 
Various sensors including 3-D Range Sensors~\cite{Range2016}, Near-Infrared Cameras~\cite{LeeCFH15}, Stereo Cameras~\cite{Stereo2009}, CCD Cameras~\cite{Qiao2008} and a combination of them~\cite{KrotoskyT07} have been employed. In what follows, we mainly review approaches with convolutional neural networks (CNNs) and models about occlusion handing.

%-------------------------------------------------------------------------
%-------------------------------------------------------------------------
\begin{figure*}[tbp]
    \centering
    \includegraphics[width=1\linewidth]{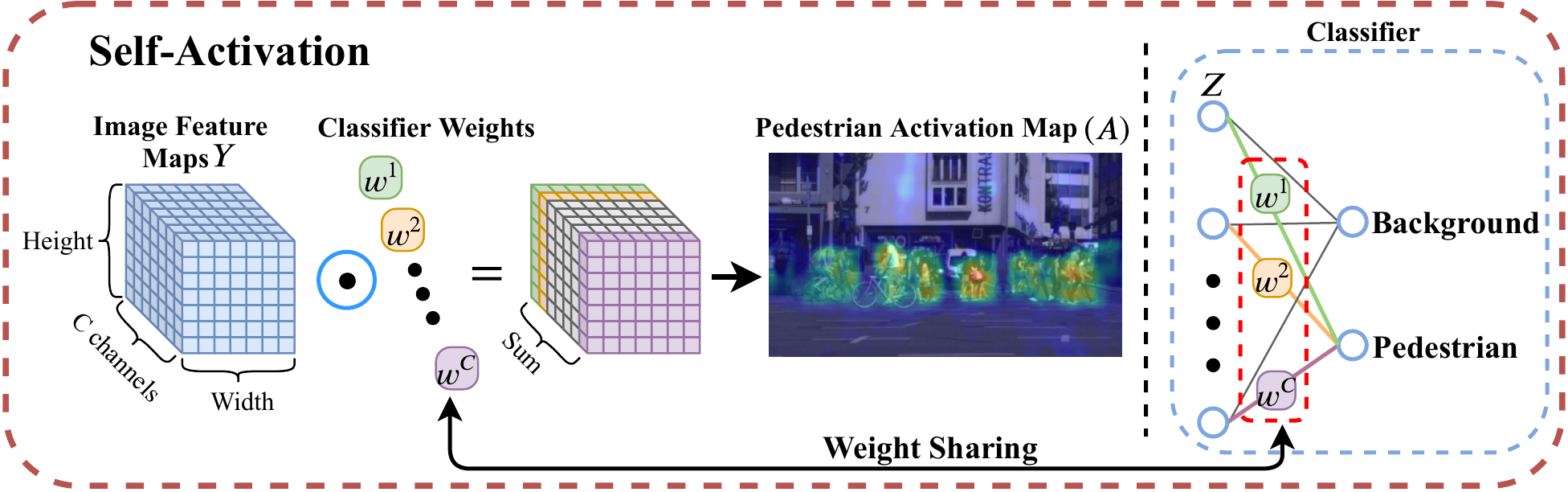}
    \caption{Self-activation module. (Best viewed in color)}
    \label{fig_sc}
\end{figure*}

\subsection{Pedestrian detection}
With the rise of deep learning, pedestrian detection methods move from hand-crafted features to CNN-based features. Early approaches focus on exploring effective network architectures and hyper-parameters for feature learning~\cite{DBLP:conf/cvpr/ZhangBOHS16}. Since 2014, RCNN~\cite{RCNN16}, which integrates high-quality region proposals with deep feature representation, has been leading the object detection area. In the following years, Fast R-CNN~\cite{Fast-RCNN2015} and Faster R-CNN~\cite{DBLP:conf/nips/RenHGS15} were proposed to aggregate feature representation and improve the detection efficiency. By using deep learning features \cite{DBLP:conf/cvpr/BellZBG16, DBLP:conf/cvpr/LinDGHHB17, Wan-MinEntropyLatentMode, Wan-CMIL} for general object detection, these approaches have achieved unprecedented good performance. In~\cite{DBLP:conf/eccv/ZhangLLH16}, Zhang \textit{et al.} borrowed the Faster R-CNN framework for pedestrian detection, by increasing the resolution of feature maps and adding hard negative mining modules.

Despite of the effectiveness of these approaches on general object detection, detecting heavily occluded pedestrians remains an open and challenging problem, as indicated by the low performance of existing state-of-the-art approaches (the miss rate is often higher than 50\% when the false positive rate per image is 0.1 \cite{DBLP:conf/iccv/BrazilYL17}). The primary reason for the low performance lies in that the occluded parts of pedestrians generate random features which can significantly decrease the representation capability of convolutional features. The problem about how to suppress the features from occluded regions while reinforcing those from visible parts of pedestrians requires to be further investigated.

%-------------------------------------------------------------------------
\subsection{Occlusion handling}
\textbf{Part-based models.} One major line of methods for occluded pedestrian detection resorts to the part-based model \cite{DBLP:conf/iccv/TianLWT15,MGTBM2013,XuVLMP14,PedersoliGHR14,LiuYDCYZ15}, which leverages a divide-and-conquer strategy, $i.e.,$ using different part detectors, to handle pedestrians with different occlusions.

In \cite{DBLP:conf/iccv/MathiasBTG13}, the Franken-classifiers learns a set of detectors, where each detector accounts for a specific type of occlusion. Zhou \textit{et al.} \cite{DBLP:conf/iccv/ZhouY17} proposed using multi-label classifiers, implemented by two fully connected layers, to localize the full body and the visible parts of a pedestrian, respectively. Zhang \textit{et al.} \cite{zhang2019circlenet} proposed CircleNet to implement reciprocating feature adaptation and used an instance decomposition training strategy.
In \cite{DBLP:conf/iccv/OuyangW13}, a joint deep learning framework was proposed and multi-level part detection maps were used for estimating occluded patterns. In \cite{DBLP:conf/eccv/ZhangWBLL18}, an occlusion-aware R-CNN (OR-CNN) was presented, with an aggregation loss and a part occlusion-aware region of interest (PORoI) pooling. The authors enforced proposals to be close to the corresponding objects, while integrating the prior structure of human body to predict visible parts.

Although effective, part-based models suffer complex occlusions as the limited number of parts experiences difficulty in covering various occlusion situations. Increasing the number of parts could alleviate such a problem but will increase the model complexity and the computational cost significantly.

%-------------------------------------------------------------------------
\textbf{Attention models.}
The other line of methods involves attention-based models~\cite{zhang2018occluded}, which replace ``hard" object parts with ``soft" attention regions by introducing attention or saliency modules \cite{wang2017repulsion,DBLP:conf/eccv/ZhangWBLL18}.

In \cite{zhang2018occluded}, the Faster R-CNN with attention guidance (FasterRCNN-ATT) was proposed to detect occluded instances. Assuming that each occlusion pattern can be formulated as a combination of body parts, a part attention mechanism was proposed to represent various occlusion patterns by squeezing the features from multiple channels.
In \cite{DBLP:journals/CVPR19_Thermal_Images}, the feature learning procedure for pedestrian detection was reinforced with pixel-wise contextual attention based on a saliency network. In \cite{SmallScaleEccv2018-Graininess}, a scale-aware pedestrian attention module was proposed to guide the detector to focus on pedestrian regions. The scale-aware attention module targets at exploiting fine-grained details in proper scales into deep convolutional features for pedestrian representation. Thermal images are used to detect pedestrians at night, but not suitable for applications in daytime. Ghose \textit{et al.} \cite{DBLP:journals/CVPR19_Thermal_Images} used saliency maps to augment thermal images, which is an attention mechanism for pedestrian detectors especially during daytime.

The introduction of attention/saliency has boosted the performance of pedestrian detection. Nevertheless, most existing approaches ignore the class-specific confidence produced by the detection network, and therefore experience difficulty in discriminating the attention regions of positives from those of negatives. In \cite{wang2017repulsion}, a repulsion loss (RepLoss) approach was designed to enforce pedestrian localization in crowded scenes. With RepLoss, each proposal is forced to be attentive to its designated targets, while kept away from other ground-truth objects. Nevertheless, the discriminative capacity of features is not reinforced despite that the spatial localization is aggregated.

\textbf{Generative models.}
Generative methods \cite{DBLP:conf/cvpr/WangGL019, DBLP:conf/iccv/QiuWT019, ouyang2018pedestrian} have been explored to produce training samples and solve the occlusion problem. The Cycle GAN \cite{DBLP:conf/cvpr/WangGL019} method transforms synthetic images to real-world scenes for data augmentation. Pedestrian-Synthesis-GAN \cite{ouyang2018pedestrian} generates labeled pedestrian data and adopts such data to enforce the performance of pedestrian detectors. Meanwhile structural context descriptor \cite{DBLP:conf/iccv/QiuWT019} is used to characterize the structural properties of individuals in crowd scenes. A weighted majority voting method \cite{DBLP:journals/tnn/Tao00YW19} inspired by domain adaptation is used to generate labels for other visual tasks.

In this paper, we propose the self-activation approach to explore the class-specific confidence predicted by the detection network. Our approach not only discriminates occluded regions from visible pedestrian parts, but also couples with the detection network to reinforce feature learning in a self-paced manner. The self-activation approach is inspired by class-activation maps (CAMs)~\cite{DBLP:conf/cvpr/ZhouKLOT16}, a kind of top-down feature activation approach. However, it is essentially different from CAMs as the activation is performed during the feature learning procedure, while that of CAM is performed after the network training is completed. Our work is also related to the squeeze-and-excitation (SE) network~\cite{SE-Net2018}, which adaptively recalibrates channel-wise feature responses by explicitly modelling inter-dependencies between channels. The difference lies in that our approach leverages the semantic information ``squeezed" in the classifier and therefore enforces the discriminative capacity of features more effectively.

%-------------------------------------------------------------------------
\section{Feature Calibration}

The core of our Feature Calibrating Network (FC-Net) is a self-activation (SA) module, as shown in Fig.\ \ref{fig_sc}, which estimates a pedestrian activation map by reusing classifier weights, without any additional parameter involved.
The pedestrian activation map is used to manipulate the network with a feature calibration (FC) module in pixel-wised and region-based manners, as shown in Fig.\ \ref{fig_spatial} and Fig.\ \ref{fig_context} respectively. The SA and FC modules are iteratively called during the network training to enforce visible parts while suppressing occluded regions.

%-------------------------------------------------------------------------
\subsection{Self-activation (SA)}

In the Faster-RCNN framework, the pedestrian classifier output, $y(Z) = f(W^T Z+b)$, is made up of a linear model and a nonlinear function.
For the binary classification problem, the network has two weight vectors in the fully-connected layer of the classifier, one for the pedestrian and the other for the background. The weight vector for the pedestrian, denoted as  $W=(w^1,w^2,...,w^C)^T\in \mathbb{R}^C$, where $C$ is the number of feature channels, as shown in Fig.\ \ref{fig_sc}. Different feature channels represented by the feature $Z$ detect different pedestrian
parts, as shown in Fig.\ \ref{fig_PAM}. To reflect the detected parts in the output $y(Z)$, their corresponding weights must be large, and if some channels only detect background parts, their corresponding weights should be small. This means that the weights actually ``squeeze" the channel-wise semantic information for pedestrian representation.

The self-activation module (Fig.\ \ref{fig_sc}) reuses the semantic information squeezed in the classifier weights to construct a pedestrian activation map. This procedure is implemented by weighting and summing all of the convolutional feature channels.
Specifically, let $Y \in \mathbb{R}^{M \times N \times C}$ be the convolutional feature maps of an image, where $M$ and $N$ respectively denote the width and height of the feature maps. An element $A_{m,n}$ on the pedestrian activation map, $A \in \mathbb{R}^{M \times N}$, is calculated as
\begin{equation}
    A_{m,n} = \sum_{c=1}^{C}{w^c \cdot Y_{m,n}^c},
\label{Eq.SA}
\end{equation}
where $m$ and $n$ denote the 2D coordinates over the feature maps, and $c$ is the index of the feature channels.

\begin{figure*}[tbp]
    \centering
    \includegraphics[width=1\linewidth]{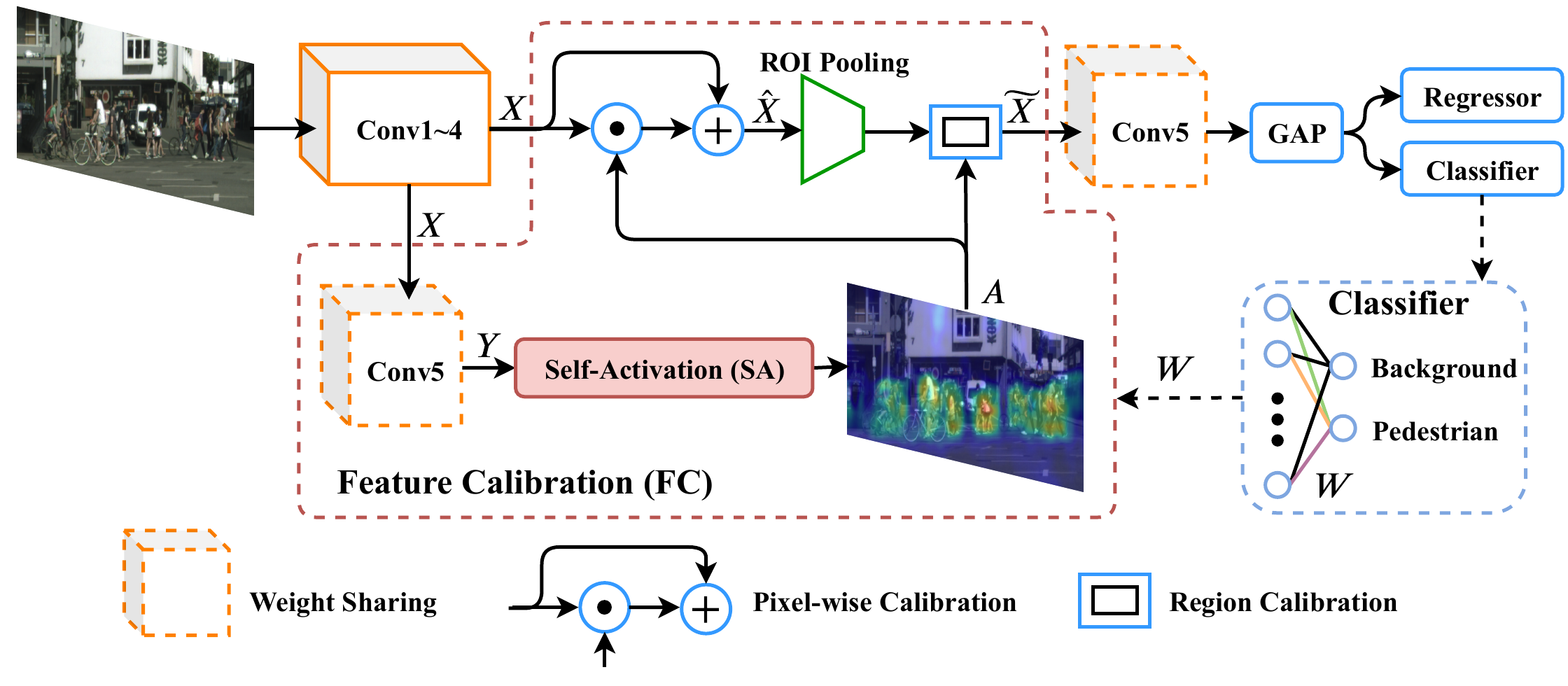}
    \caption{\label{Fig.2} The architecture of the feature calibration network (FC-Net), which is made up of a deep detection network, a self-activation (SA) module, and a feature calibration (FC) module. After each feed-forward procedure, the detection network learns the classifier weights, which are reused by the SA module for the calibration of the convolutional features. With multiple iterations of the feature calibration, FC-Net learns features which highlight the visible parts and suppress the occluded regions of pedestrians. In the network, GAP stands for global average pooling.}
    \label{fig_framework}
\end{figure*}

The baseline detector is the Faster RCNN equipped with ResNet, which has a global average pooling (GAP) layer after Conv5, as shown in Fig.\ \ref{fig_framework}. The GAP layer converts the multiple values of a feature map (channel) to a single value. As a result, multiple feature maps are converted to a vector, which has the same element number with the classifier.

For a pedestrian, different feature channels are sensitive to different parts as the convolutional filters are learned for different visual patterns ($w^c\cdot Y^c$). Benefiting from the fact that the RoI pooling (see Fig.\ \ref{fig_framework} later for the detail) does not change the order of feature channels, the learning procedure constructs a statistical relationship between the feature channels and the weight vector. The larger is a weight element, the more informative is the corresponding feature channel.
With Eq.\ \ref{Eq.SA}, we can aggregate visual patterns into a pedestrian activation map, which indicates the statistical importance of pixels for pedestrian representation. With the pedestrian activation map, we can enforce the features from visible pedestrian parts, as well as depressing occluded regions when the values of either the corresponding feature channels or the weights are small.

%-------------------------------------------------------------------------
\subsection{Feature calibration (FC)}

\begin{figure}[tbp]
    \centering
    \includegraphics[width=1.0\linewidth]{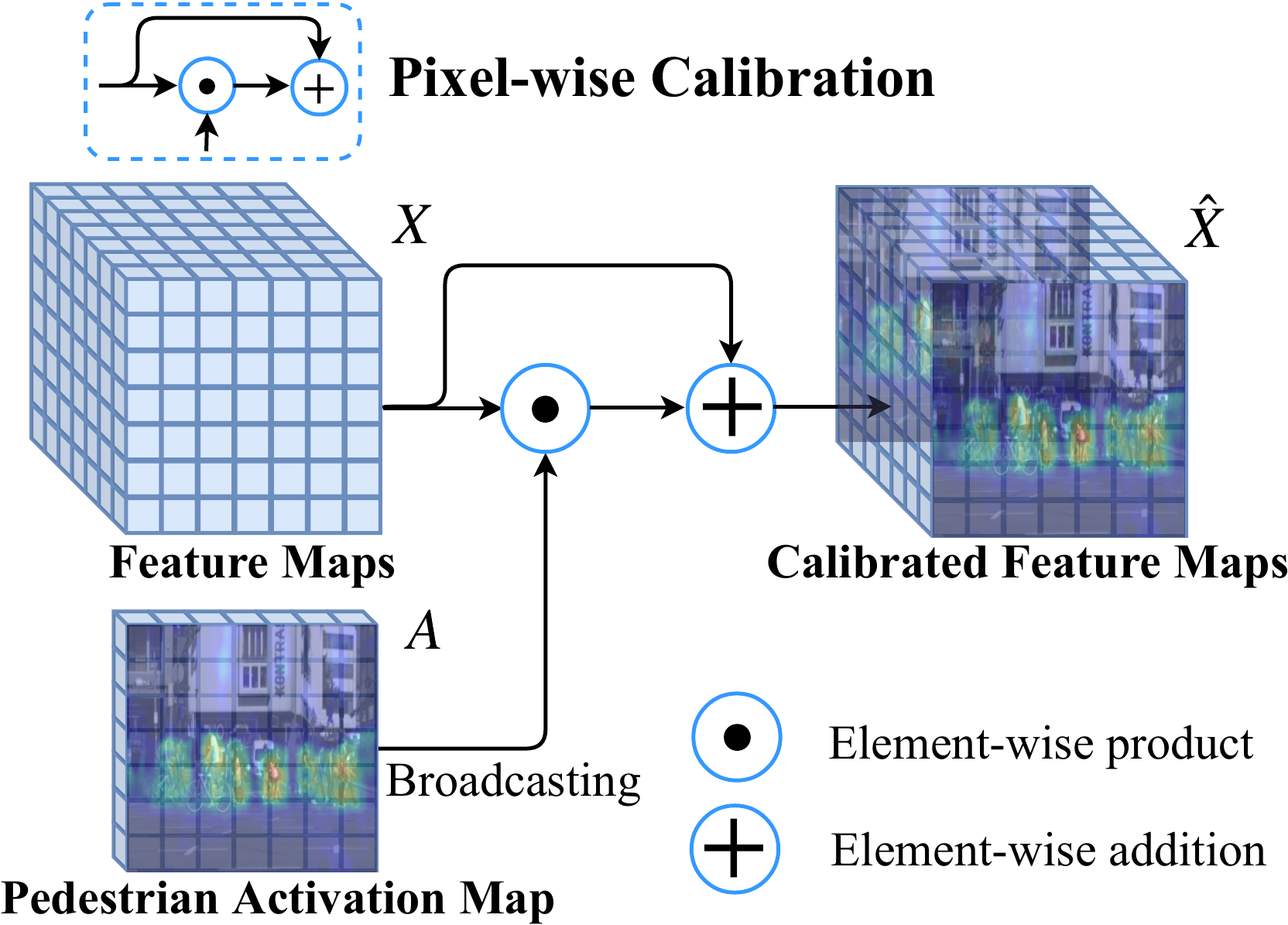}
    \caption{Pixel-wise calibration. (Best viewed in color)}
    \label{fig_spatial}
\end{figure}

To make use of the information incorporated into the pedestrian activation map, we follow it with a feature calibration step which aims to aggregate the convolutional features. Towards this goal, such calibration is expected to handle occlusion effectively.
First, it should be adaptive to spatial occlusion (in particular, it must be capable of suppressing the channels which output high feature values on occluded regions), and second, it should incorporate the context information so that when an important part of pedestrians is occluded, the region features can still be used for detection.

To meet these requirements, we design pixel-wise calibration and region-based calibration. The former enforces the feature maps to focusing on visible and discriminative parts of pedestrians, while the latter leverages the pedestrian activation map to select the most discriminative regions via introducing multi-level context information.

The pixel-wise calibration reinforces or suppresses the convolutional features in the learning procedure according to the pedestrian activation map. When an important part of pedestrians was occluded, the context regions were validated to provide discriminate information from the perspective of concurrence. For example, pedestrians often stay on the sidewalk or bicycles, but seldom in the air. The region calibration module can leverage the features of context regions for better detection.

%-------------------------------------------------------------------------
\begin{figure}[tbp]
    \centering
    \includegraphics[width=1.0\linewidth]{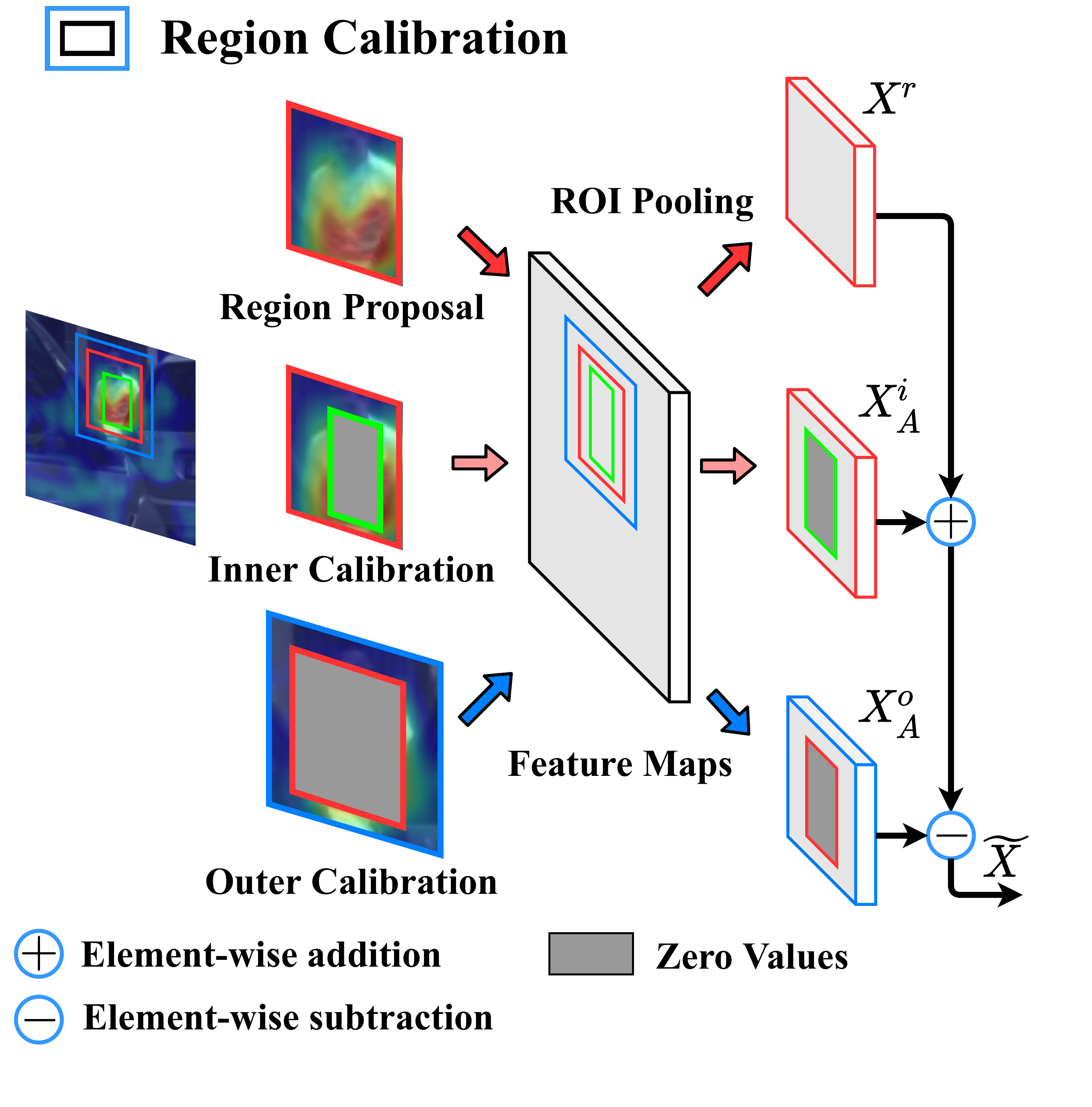}
    \caption{Region calibration.}
    \label{fig_context}
\end{figure}

%-------------------------------------------------------------------------

\textbf{Pixel-wise calibration.} The pixel-wise feature calibration, as shown in Fig.\ \ref{fig_spatial}, is performed with a pixel-wise product operation and an addition operation. Denoting the feature maps before and after the calibration as $X=\{X^c\}$ and $\hat X=\{\hat {X^c} \}$, with $c$ being the channel index, the pixel-wise calibration operation is performed as
\begin{equation}
    \hat X^c = A \odot X^c+X^c, ~c=1,...,C,
    \label{Eq.PixelwiseCal}
\end{equation}
where $\odot$ denotes the element-wise product. The calibrated features are then inserted into the network for other feature computation.

Note that the pixel-wise calibration is performed with a product and an addition operations. The product operation converts the occlusion and non-occlusion confidence reflected by the pedestrian activation map to each feature channel. Nevertheless, given pedestrians of various appearances and clutter backgrounds, the pedestrian activation map is not necessarily accurate.
The usage of the addition operation keeps the original features and thus smooths the effect of pixel-wise calibration. 
As the pedestrian activation map (PAM) is calculated by weighting and summing all of the convolutional features, it combines the discriminative information from both classifier weights and features to highlight visible pedestrian parts, in a self-activation fashion. The classifier weights themselves, however, can not indicate visible or occluded parts of pedestrians.

%-------------------------------------------------------------------------

%-------------------------------------------------------------------------
\textbf{Region calibration.}
Given the pedestrian activation map, an adaptive context module can be further developed to enhance the feature representation towards good detection.

As shown in Fig. \ref{fig_context}, we first define an inner calibration region and an outer calibration region for each region proposal. The outer calibration region is defined as a rectangle similar to the region proposal but with height $=h\times r$ and width $=w\times r$, where $h$ and $w$ are the height and width of the region proposal, respectively, and $r>1$ is a hyper-parameter. Similarly, the inner calibration region is defined as a rectangle with its height $=h/r$ and width $=w/r$. The inner calibration region is inside the region proposal and covers the area with the largest sum of pixel values on the pedestrian activation map. The outer calibration region covers the region proposal and is with the same center as the inner calibration region. The coordinates of the calibration regions are determined with an exhaustive search around the region proposal.

From the above definitions, we know that the locations of the two calibration rectangles are determined by the pedestrian activation map $A$. Let $X^r$, $X^i_A$, and $X^o_A$ be three features of the same size after RoI pooling. As shown in Fig. \ref{fig_context}, both $X^r$ and $X^i_A$ are from the region proposal, but the features of the inner calibration rectangle inside $X^i_A$ are set to 0; and $X^o_A$ is from the outer calibration rectangle, but the features of the region proposal inside $X^o_A$ are set to 0. Then the calibrated feature $\widetilde X$ of the region proposal is calculated as
\begin{equation}
\widetilde{X} = X^r + X_A^i- X_A^o.
\label{eq_context}
\end{equation}

%-------------------------------------------------------------------------
In three cases, we describe the effects of Eq. \ref{eq_context} below. (i) When the region proposal perfectly detects a pedestrian, overall, the feature values of both $X^r$ and $X_A^i$ are large, while those of $X_A^o$ are small. Then $\widetilde{X}$ is enhanced significantly. (ii) When the region proposal only covers part of a pedestrian, overall, all the feature values of $X^r$, $X_A^i$, and $X_A^o$ are relatively large, which results in little enhanced/depressed features. (iii) When a pedestrian takes only a small part of the region proposal, overall, all the feature values of $X^r$, $X_A^i$, and $X_A^o$ are relatively small, which results in no enhancement of the features. From these effects, we can see that with the guidance of the pedestrian activation map $A$, the features are calibrated towards good detection.

There are two common factors which cause missing detection of occluded pedestrians. First, the occluded parts introduce significant noises to features, which could confuse the detector towards miss-classification. Second, the features of the visible parts could be not discriminative enough to detect occluded pedestrians, particularly when the background is complex. Leveraging the pedestrian activation map as an indicator, we use the pixel-wise and region calibration modules to enhance features of the visible parts and suppress those of the occluded parts, improving the opportunity to detect occluded pedestrians. When an important part of pedestrians was occluded, the context regions were validated to provide discriminate information from the perspective of concurrence. For example, pedestrians often stay on the sidewalk or bicycles, but seldom or the air. The region calibration module can leverage the features of the context regions for better detection. 

The region calibration is fulfilled by a spatial pooling function, which aggregates the features in the context areas. With region calibration, the proposal region features, inner calibration features and outer calibration features are fused. In the procedure, the information within the context region is not removed but fused according to a negative weight, so that the outer region does not cover any pedestrian part. This facilitates improving pedestrian localization accuracy.

%-------------------------------------------------------------------------
\section{Network Structure}

Based on the Faster-RCNN framework and the proposed SA and FC modules, we construct the pedestrian object detector, FC-Net, as shown in Fig.\ \ref{fig_framework}. Following the last convolutional layer (Conv5) in the detection network, the convolutional features of each region proposal are spatially pooled into a $C$-dimensional feature vector with a global average pooling (GAP) layer, where $C$ denotes the number of the feature channels. Such a feature vector is then converted into the confidence for a class of object (pedestrian or background), by multiplying it and the weight vector of the fully connected layer with a soft-max operation.

With the feature calibration and network learning, FC-Net works in this way: $X\rightarrow W\rightarrow A \rightarrow \hat X \cdots$. During the learning procedure of the network, the features of various pedestrian instances are aggregated to the classifier weight vector $W$. With the SA and FC modules, the weight vector is employed to generate the activation map $A$, which is further used to reinforce or depress the features $X$. With multiple iterations of learning, FC-Net actually implements a special kind of self-paced feature learning. The SA and FC modules are stacked together to form a new architecture, which is universal for deep learning-based object detection.

The proposed SA and FC modules are extremely compressed without additional parameters, involving channel-wise feature calibration, $i.e.$, loosely speaking, $\hat X = A \odot X$ and $A=W\cdot X$, where $W$ is borrowed from the detection network. $W$ involves in the forward process and improves the feature learning process of FC-Net, by fully investigating the high-level semantic information ``squeezed" in the classifier. During the feature calibration procedure, the semantic information is excited to activate the feature maps so that they can focus on visible pedestrian parts while suppressing occluded regions.

%-------------------------------------------------------------------------
\section{Experiments}

%-------------------------------------------------------------------------
\begin{figure*}[!tbp]
    \subfloat[Baseline]{\label{Fig.activation_map_SUM}
    \begin{minipage}{1.0\linewidth}
        \centering
        \includegraphics[width=18.0cm]{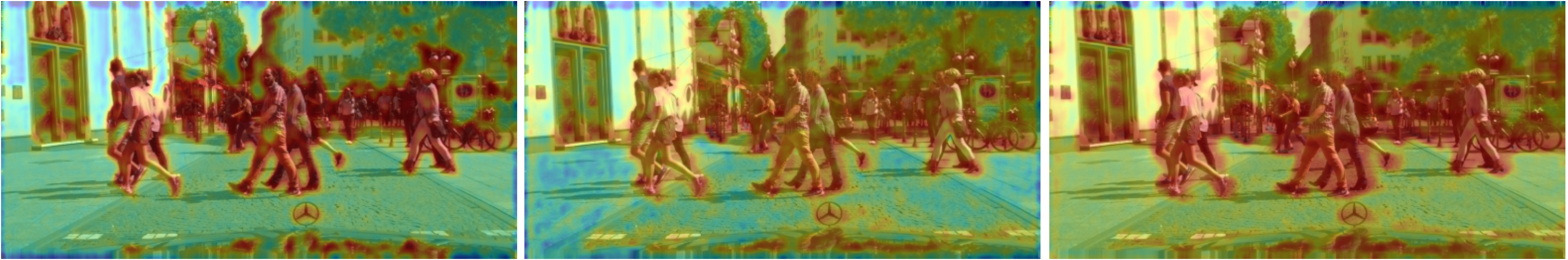}
    \end{minipage}}

    \subfloat[Squeeze-and-excitation]{\label{Fig.activation_map_SE}
    \begin{minipage}{1.0\linewidth}
        \centering
        \includegraphics[width=18.0cm]{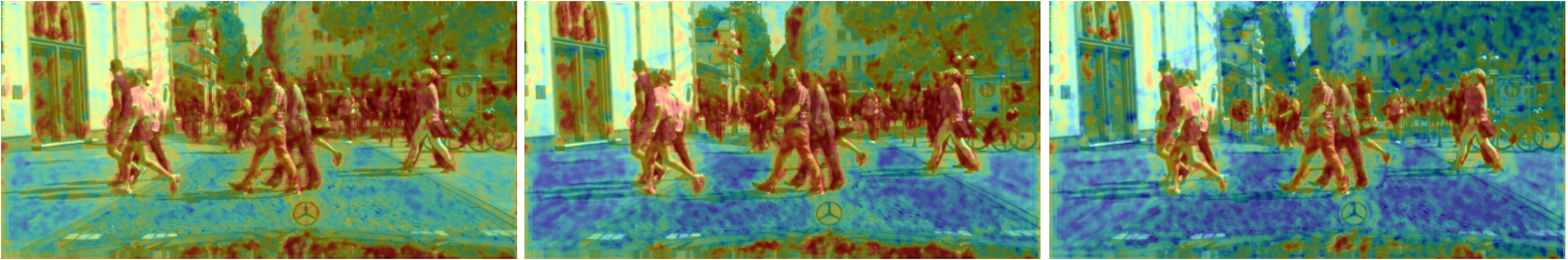}
    \end{minipage}}

    \subfloat[Self-Activation]{\label{Fig.activation_map_ours}
    \begin{minipage}{1.0\linewidth}
        \centering
        \includegraphics[width=18.0cm]{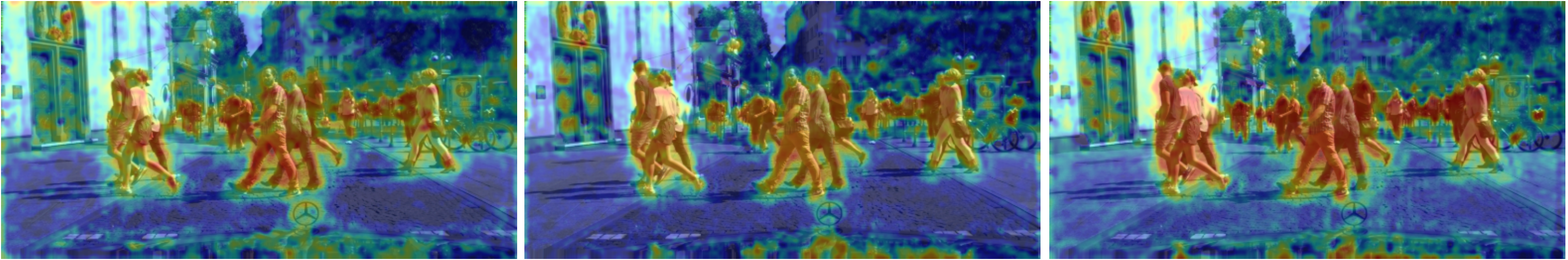}
    \end{minipage}}
    \caption{\label{Fig.Exp_SA} Comparison of pedestrian activation maps. The proposed SA module can more effectively highlight the visible parts of the pedestrians than the squeeze-and-excitation (SE) network \cite{SE-Net2018}. (Best viewed in color)}
\end{figure*}

In this section, we first describe the experimental settings about datasets, evaluation metrics, and implementation details. We then evaluate the effectiveness of the proposed SA and FC modules on the benchmark datasets. Finally, the performance of FC-Net and the comparisons with state-of-the-art pedestrian detectors are presented.

\subsection{Experimental settings}
\label{sec:Experimental_Settings}
\textbf{Datasets:} Two common datasets, Caltech \cite{DBLP:conf/cvpr/DollarWSP09}  and CityPersons \cite{DBLP:conf/cvpr/ZhangBS17}, are used to evaluate FC-Net.
The Caltech dataset contains approximately 10 hours of street-view videos taken with a camera mounted on a vehicle. The most challenging aspect of the dataset is the large number of low-resolution and occluded pedestrians. We sample 42,782 images from set00 to set05 for training and 4,024 images from set06 to set10 for testing.
The CityPersons dataset is built upon the semantic segmentation dataset Cityscapes \cite{DBLP:conf/cvpr/CordtsORREBFRS16}. It contains 18 different cities in Germany in three different seasons and various weather conditions. There are 5,000 images, 2,975 for training, 500 for validation, and 1,525 for testing. This dataset is much more ``crowded" than Caltech, and the most challenging aspect of the pedestrian objects is heavy occlusion.

\textbf{Evaluation Metric:}
To demonstrate the effectiveness of FC-Net under various occlusion levels, we follow the strategy in \cite{wang2017repulsion} and \cite{DBLP:conf/eccv/ZhangWBLL18} to define three subsets from the validation set in CityPersons: (i) \textit{Reasonable} (occlusion $<$ $35\%$ and height $>$ $50$ pixels), (ii) \textit{Partial} ($10\%$ $<$ occlusion $<$ $35\%$ and height $>$ $50$ pixels), and (iii) \textit{Heavy} (occlusion $>$ $35\%$ and height $>$ $50$ pixels). The commonly used average-log miss rate $\mathbf{MR}^{-2}$ computed in the False Positive Per Image (FPPI) range of [$10^{-2}, 10^{0}$] \cite{DBLP:conf/cvpr/ZhangBS17} is used as the performance metric.
%-------------------------------------------------------------------------

\textbf{Implementation details:} The baseline detection network is the commonly used Faster R-CNN \cite{DBLP:conf/nips/RenHGS15}.
It is specified for pedestrian detection by following the settings in \cite{DBLP:conf/cvpr/ZhangBS17}. ResNet-50 \cite{DBLP:conf/cvpr/HeZRS16} is used as the backbone network as it is faster and lighter than VGG-16. By using Faster R-CNN as the baseline detection network, we achieve 15.18$\%$ $\mathbf{MR}^{-2}$ on the  CityPersons validation set, which is sightly better than the reported result, 15.4$\%$ $\mathbf{MR}^{-2}$, in \cite{DBLP:conf/cvpr/ZhangBS17}.

The implementation details of FC-Net are consistent with that of the maskrcnn-benchmark project \cite{massa2018mrcnn}. We train the network for 6k iterations, with the base learning rate set to 0.008 and decreased by a factor of 10 after 5k iterations on CityPersons. The Stochastic Gradient Descent (SGD) solver is adopted to optimize the network on 8 Nvidia V100 GPUs. A mini-batch involves 1 image per GPU. The weight decay and momentum are set to 0.0001 and 0.9, respectively. We only use single-scale training and testing samples ($\times$1 or $\times$1.3) for fair comparisons with other approaches.

%-------------------------------------------------------------------------
\begin{table*}[]
\centering
\fontsize{9}{12}\selectfont
\begin{tabular}{c|c|c|c|c|c|c|c|c|c}
\hline
\multicolumn{3}{c|}{Method}                                             & Scale       & \multicolumn{2}{c|}{\textit{Reasonable}} & \multicolumn{2}{c|}{\textit{Heavy}} & \multicolumn{2}{c}{\textit{Reasonable+Heavy}} \\ \hline \hline
\multirow{7}{*}{FC-Net} & Pixel-wise calibration & Region calibration &             & MR               & $\Delta$MR   & MR                & $\Delta$MR   & MR                 & $\Delta$MR    \\ \cline{2-10}
                        &                        &                      & $\times$1   & 15.18            & -            & 51.05             & -            & 31.05              & -             \\ \cline{2-10}
                        & $\surd$                & $\surd$              & $\times$1   & 13.93            & +1.25        & 46.79             & +4.26        & 29.64              & +1.41         \\ \cline{2-10}
                        &                        &                      & $\times$1.3 & 13.21            & -            & 46.75             & -            & 29.45              & -             \\ \cline{2-10}
                        & $\surd$                &                      & $\times$1.3 & 12.43            & +0.78        & 43.60             & +3.15        & 27.80               & +1.65         \\ \cline{2-10}
                        &                        & $\surd$              & $\times$1.3 & 11.71            & +1.50        & \textbf{41.36}    & +5.39        & 26.70               & +2.75          \\ \cline{2-10}
                        & $\surd$                & $\surd$              & $\times$1.3 & \textbf{11.63}   & +1.58        & 42.77             & +3.98        & \textbf{26.21}     & +3.24         \\ \hline
\end{tabular}
\caption{\label{Table_Validation_Results} Ablation study of the proposed feature calibration (FC) module on the CityPersons validation dataset with $\mathrm{MR}^{-2}$. Smaller numbers indicates better performance.}
\end{table*}

%-------------------------------------------------------------------------
\begin{figure}[tbp]
    \centering
    \includegraphics[width=1.0\linewidth]{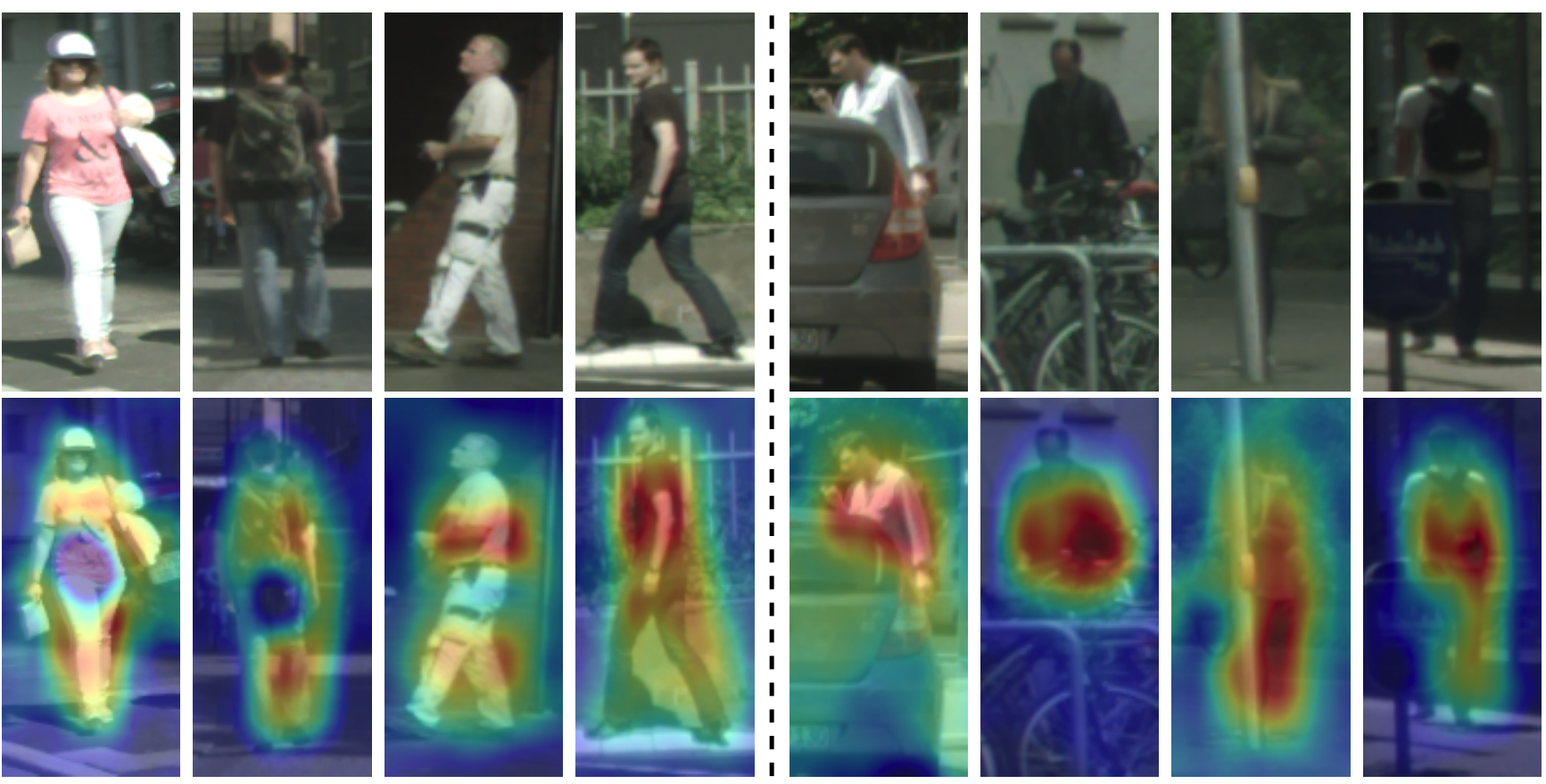}
    \caption{Examples of pedestrians and pedestrian activation maps. Left: non-occluded instances. Right: occluded instances.}
    \label{fig_PAM_Pedestrian}
\end{figure}

%-------------------------------------------------------------------------
\begin{figure}[tbp]
    \centering
    \includegraphics[width=1\linewidth]{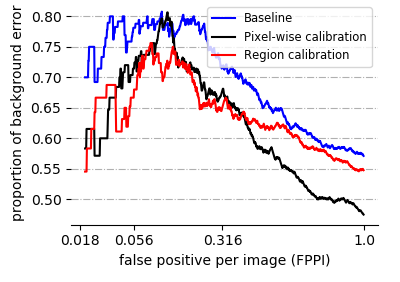}
    \caption{\label{SA_background_error} By applying the proposed feature calibration (FC) module, the proportion of false positives caused by background is significantly reduced. }
\end{figure}

% -----------------------------------------------------------------------------------
\subsection{Self-activation (SA)}
\label{sec:Ablation-Experiments}

For feature activation, squeeze-and-excitation (SE) is one of the most related works based on the self-attention mechanism to calibrate convolutional features \cite{SE-Net2018}.
Our SA module clearly differs from it, without any additional parameter added, and only reuses the weight vector of the classifier to enhance feature learning. As shown in Fig.\ \ref{Fig.Exp_SA}, our SA module can more effectively highlight the visible parts of the pedestrians than the SE network~\cite{SE-Net2018}. This shows that the high-level semantic information is crucial to suppress the backgrounds and enhance the foregrounds.

Fig.\ \ref{Fig.activation_map_ours} compares the results of different methods from top to bottom rows. From left to right, we show the results of the same methods at different iterations. Compared with the baseline Faster RCNN \cite{DBLP:conf/nips/RenHGS15-s} (first row) and the SE network \cite{SE-Net2018} (second row), the proposed SA module effectively highlights the pedestrian regions while depressing the background.

As the weights squeeze the statistical importance of pedestrian parts and feature channels, they are used to aggregate the parts/channels into an activation map, which enforces visible parts while depressing occluded parts. Fig.\ \ref{fig_PAM_Pedestrian} shows the pedestrian activation maps of some non-occluded instances and occluded instances. It can be seen that our approach can adaptively suppress various occluded regions and enforce visible parts of pedestrian objects.
%-------------------------------------------------------------------------
\begin{table}[tbp]
\centering
\fontsize{9}{12}\selectfont
\begin{tabular}{c|c|c|c|c|c}
\hline
Ratio (Height) & 1.0   & 1.4  & 1.6   & 1.8            & 2.0   \\ \hline \hline
\textit{Reasonable}     & 13.21 & 13.00 & 12.40 & \textbf{11.71} & 12.78 \\ \hline
\textit{Heavy}    & 46.75 & 43.34 & 42.10 & \textbf{41.36} & 42.09 \\ \hline
\textit{Reasonable+Heavy}  & 29.45 & 27.75 & 27.13 & \textbf{26.70}  & 27.11 \\ \hline
\end{tabular}
\caption{\label{Table_Different_Ratios_Height} With the width ratio $= 1$, $\mathrm{MR}^{-2}$ under different ratios for height between the outer rectangle and the region proposal, and between the region proposal and the inner rectangle (see Fig.\ \ref{fig_context}).}
\end{table}

\begin{table}[tbp]
\centering
\fontsize{9}{12}\selectfont
\begin{tabular}{c|c|c|c|c}
\hline
Ratio (Width) & 1.0      &     1.4 &     1.6    &     1.8                 \\ \hline \hline
\textit{Reasonable}            & \textbf{11.71} & 12.09     & 12.90      & 12.87                      \\ \hline
\textit{Heavy}           & \textbf{41.36} & 42.51     & 43.29      & 42.98                      \\ \hline
\textit{Reasonable+Heavy}         & \textbf{26.70}  & 27.53     & 28.23      & \multicolumn{1}{c}{28.81} \\ \hline
\end{tabular}
\caption{\label{Table_Different_Ratios_HW} With the height ratio $= 1.8$, $\mathrm{MR}^{-2}$ under different ratios for width between the outer rectangle and the region proposal, and between the region proposal and the inner rectangle (see Fig.\ \ref{fig_context}).}
\end{table}

%-------------------------------------------------------------------------
\subsection{Feature calibration (FC)}

\textbf{Pixel-wise calibration:} 
In pedestrian detection, \textit{background} is an important factor causing detection errors \cite{DBLP:conf/cvpr/ZhangBOHS16}, \cite{wang2017repulsion}. We therefore propose using background error to validate the effect of the FC module. One background error is defined as the case when the intersection over union (IoU) between a detection result and the ground truth is less than 0.2. Fig.\ \ref{SA_background_error} compares the error from background before and after using feature calibration. It can be seen that our pixel-wise calibration effectively reduces missed detections and false positives caused by background noise. In Fig.\ \ref{SA_background_error}, the blue curve shows the background errors of the baseline are very significant, i.e., larger than $70\%$ from $FPPI=0.056$ to $FPPI=0.316$. By using our pixel-wise calibration module, the background errors are significantly reduced (black curve), especially from $FPPI=0.316$ to $FPPI=1.0$. At $FPPI=1.0$, FC-Net with the pixel-wise calibration reduces background errors from $58\%$ to $48\%$. This shows that with the pixel-wise feature calibration the background errors are significantly suppressed.

In Table \ref{Table_Validation_Results}, we quantitatively evaluate the effect of the pixel-wise feature calibration. Compared with the baseline, FC-Net with the pixel-wise feature calibration reduces $0.78\%$ MR$^{-2}$ at $\times$1.3 scale on the \textit{Reasonable} subset, $3.15\%$ MR$^{-2}$ on the \textit{Heavy} subset, and $1.65\%$ MR$^{-2}$ on the \textit{Reasonable+Heavy} subset.
%-------------------------------------------------------------------------
\begin{figure}[tbp]
    \centering
    \subfloat[]{\label{Fig.false_postives}
    \begin{minipage}{1.0\linewidth}
        \includegraphics[width=8.5cm]{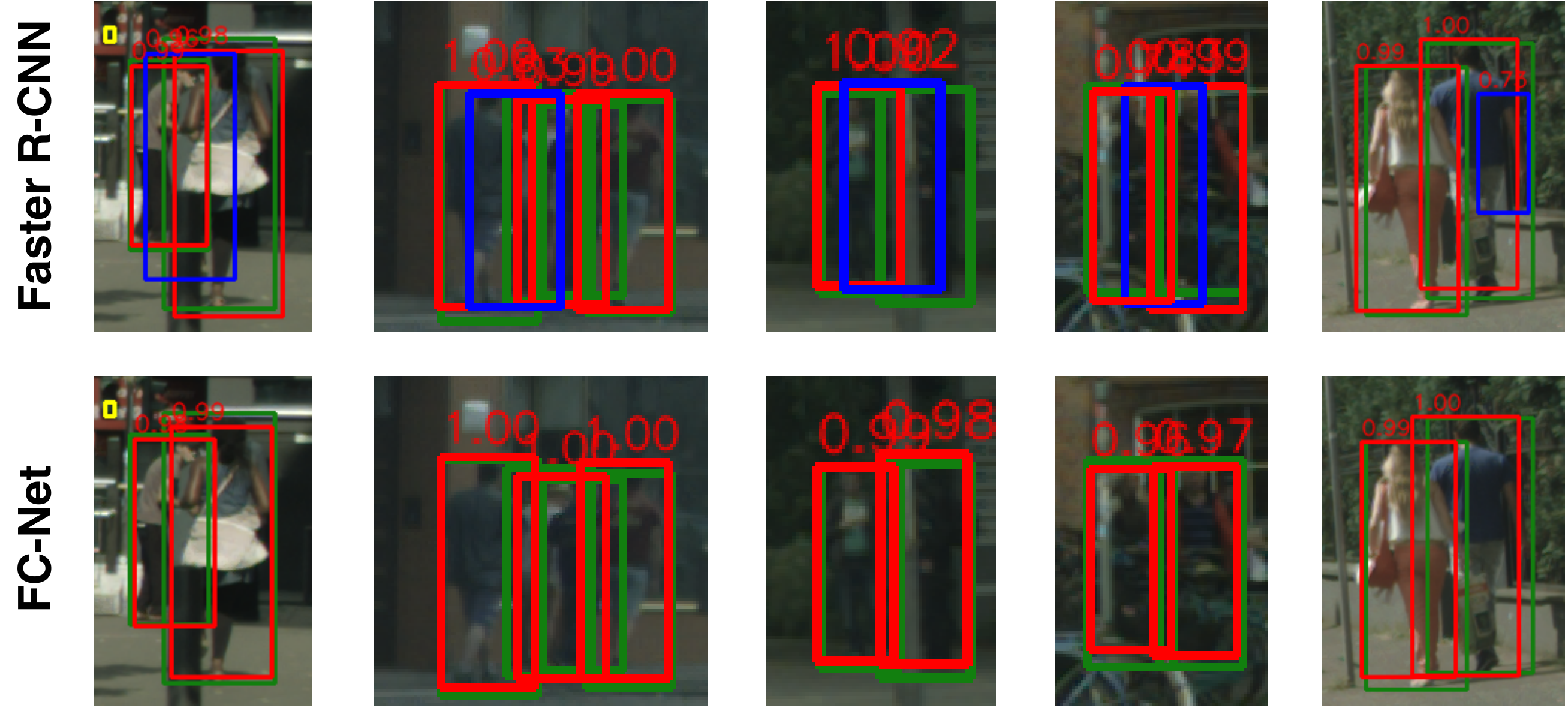}
    \end{minipage}}

    \centering
    \subfloat[]{\label{Fig.missed_detections}
    \begin{minipage}{1.0\linewidth}
        \includegraphics[width=8.5cm]{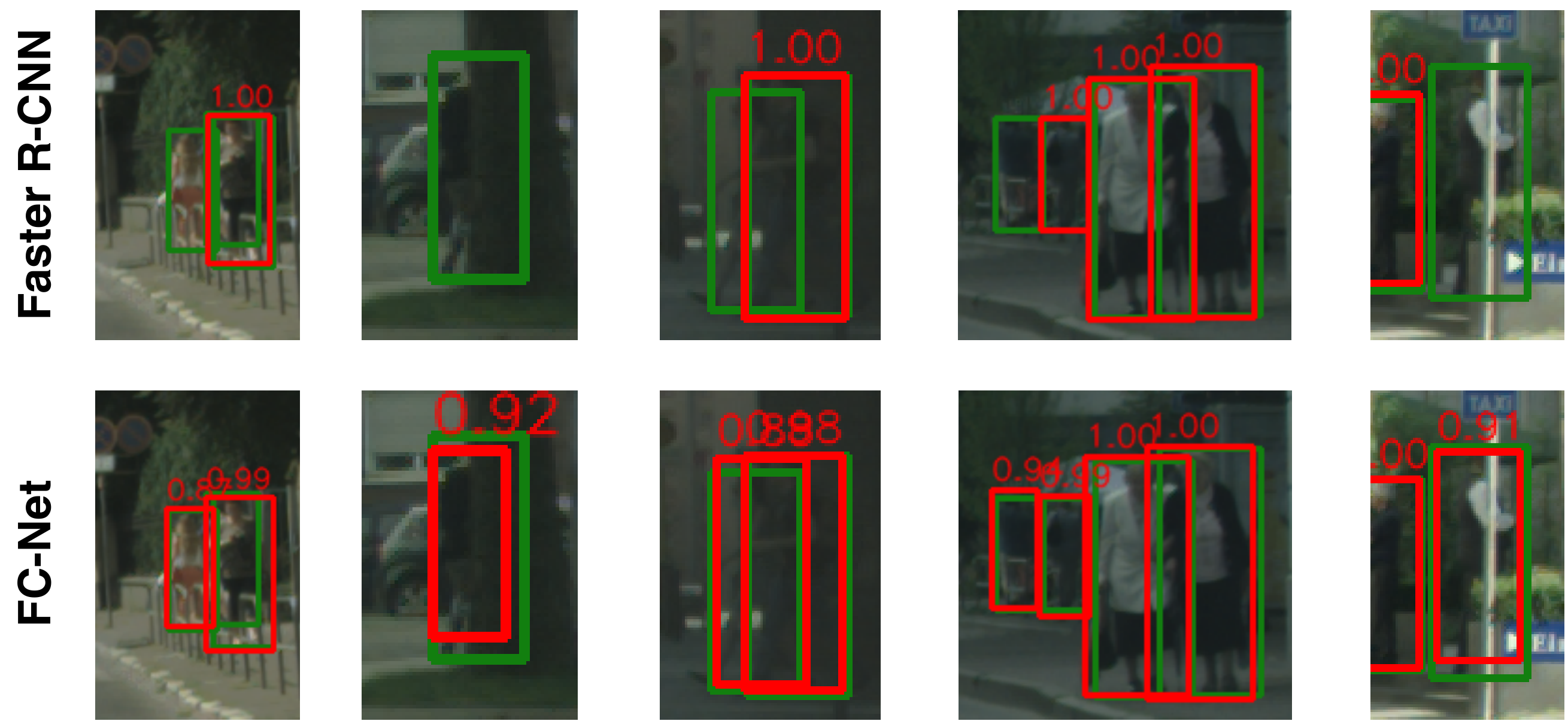}
     \end{minipage}}
    \caption{Comparison of Faster R-CNN and FC-Net on occluded pedestrians. The red bounding boxes indicate correctly detected pedestrians. The blue boxes indicate false positives and the green boxes the ground-truth. (a) FC-Net produces fewer false positives. (b) FC-Net detects more occluded pedestrians than Faster R-CNN.}
    \label{Fig.occlusion_comparison}
\end{figure}

% -----------------------------------------------------------------------------------
\textbf{Region calibration:}  By using the region calibration module, the background errors are significantly reduced (red curve in Fig.\ \ref{SA_background_error}). At FPPI=0.056, FC-Net with the region calibration reduces the proportion of the background errors from 76\% to 66\%.

In Table \ref{Table_Validation_Results}, FC-Net with the region calibration reduces
1.50$\%$ MR$^{-2}$ at $\times$1.3 scale on the \textit{Reasonable} subset,
5.39$\%$ MR$^{-2}$ on the \textit{Heavy} subset, and 2.75$\%$ MR$^{-2}$ on the \textit{Reasonable+Heavy} subset,

%-------------------------------------------------------------------------
\begin{table}[!t]
\centering
\fontsize{8}{10}\selectfont
\begin{tabular}{c|c|c|c|c}
\hline
Method                               & Scale       & \textit{Reasonable}    & \textit{Heavy} & \textit{Partial} \\ \hline \hline
\multirow{2}{*}{Adapted FasterRCNN \cite{DBLP:conf/cvpr/ZhangBS17}} & $\times$1   & 15.4          & -                 & -                   \\ \cline{2-5}
                                     & $\times$1.3 & 12.8          & -                 & -                   \\ \hline
\multirow{2}{*}{Repulsion Loss \cite{wang2017repulsion}}      & $\times$1   & 13.2          & 56.9              & 16.8                \\ \cline{2-5}
                                     & $\times$1.3 & 11.6        & 55.3              & 14.8                \\ \hline
\multirow{2}{*}{OR-CNN \cite{DBLP:conf/eccv/ZhangWBLL18}}              & $\times$1   & 12.8          & 55.7              & 15.3                \\ \cline{2-5}
                                     & $\times$1.3 & \textbf{11.0} & 51.3             & {13.7}                \\ \hline
\multirow{2}{*}{CircleNet \cite{zhang2019circlenet}}              & $\times$1   & -          & -              & -                \\ \cline{2-5}
                                     & $\times$1.3 & 11.8 & 50.2             & {12.2}                \\ \hline
\multirow{2}{*}{AEMS-RPN \cite{wang2019fast}}              & $\times$1   & 13.7          & -              & -                \\ \cline{2-5}
                                     & $\times$1.3 & 12.2 & -             & -                \\ \hline

\multirow{2}{*}{FC-Net (Ours)}              & $\times$1   & 13.5          & 44.3              & 14.0                \\ \cline{2-5}
                                     & $\times$1.3 & 11.6          & \textbf{42.8}    & \textbf{11.9}       \\ \hline
\end{tabular}
\caption{\label{Table_Compared_With_SOTA} Comparison with the state-of-the-art methods on the CityPersons validation set with $\mathrm{MR}^{-2}$.}
\end{table}
%-------------------------------------------------------------------------
\begin{table*}[]
\centering
\fontsize{9}{11}\selectfont
\begin{tabular}{c|c|c|c|c|c}
\hline
Method             & Scale        & \textit{All} & \textit{Reasonable} & \textit{Reasonable\_small} & \textit{Heavy} \\ \hline \hline
Adapted FasterRCNN \cite{DBLP:conf/cvpr/ZhangBS17} & $\times 1.3$ & 43.86        & 12.97               & 37.24                      & 50.47                          \\ \hline
Repulsion Loss \cite{wang2017repulsion}    & $\times 1.5$ & \textbf{39.17}        & 11.48               & 15.67                      & 52.59                          \\ \hline
OR-CNN \cite{DBLP:conf/eccv/ZhangWBLL18}             & $\times 1.3$ & 40.19        & \textbf{11.32}               & \textbf{14.19}                      & 51.43                          \\ \hline
FC-Net (Ours)            & $\times 1.3$ & 39.26        & 12.24               & 16.67                      & \textbf{41.14}                      \\ \hline
\end{tabular}
\caption{Comparison with the state-of-the-art methods on the CityPersons test dataset with  $\mathrm{MR}^{-2}$. The results of our approach are evaluated by the authors of CityPersons and the compared results are from the official website of CityPersons\protect\footnotemark.}
\label{Table_CityPersons_Test}
\end{table*}
\footnotetext{https://bitbucket.org/shanshanzhang/citypersons.}

The positions of the inner and outer calibration rectangles are determined by searching the regions of highest value sum on the activation map. What we need to determine is the ratio parameter $r$ empirically. As shown in Table\ \ref{Table_Different_Ratios_Height}, by searching in the range [1.0, 2.0], we observe that the best performance is achieved at $r=1.8$ for height. With the best ratio for height, we further observe that the best width ratio\footnote{Note that the ratio for width may be different from that for height.} is 1.0 as shown in Table\ \ref{Table_Different_Ratios_HW}. 
The reason of the width ratio smaller than the height ratio is that other pedestrians in horizontal directions may exist in the surrounding regions of a proposal, which may confuse the detector. Note that the pixel-wise calibration does not rely on any context information and therefore is effective even in crowded scenes.

The width ratio and the height ratio are two hyper-parameters for region calibration. The ablation experiments in Table\ \ref{Table_Different_Ratios_Height} and Table\ \ref{Table_Different_Ratios_HW} show that the context information in the vertical direction is more important than that in the horizontal direction. The reason could be that there is more concurrence information between pedestrians and backgrounds in the vertical direction. When there are multiple pedestrians in horizontal directions, the context information could be interfered.

\subsection{Occlusion handling}

To show the effectiveness of the proposed SA and FC modules for occlusion handling, we evaluate the detection performance on the validation set of CityPersons where there exist significant person-to-person and car-to-person occlusions.

In Fig.\ \ref{Fig.occlusion_comparison}, we compare the detection results of Faster R-CNN and FC-Net on occluded samples. It can be seen that FC-Net produces fewer false positives and detects more pedestrians than Faster R-CNN.

%-------------------------------------------------------------------------
\begin{table}[]
\centering
\fontsize{7.6}{10}\selectfont
\begin{tabular}{c|c|c|c}
\hline
Method              & \textit{Reasonable} & \textit{Heavy} & \textit{Reasonable+Heavy} \\ \hline \hline
FasterRCNN+ATT \cite{zhang2018occluded} & 15.96      & 56.66       & 38.23         \\ \hline
FC-Net+ATT         & 14.82      & 49.02       & 31.01         \\ \hline
FC-Net (Ours)       & \textbf{13.93} & \textbf{46.79} & \textbf{29.64}  \\ \hline
\end{tabular}
\caption{Comparison with the state-of-the-art FasterRCNN+ATT \cite{zhang2018occluded} on the CityPersons validation set with $\mathrm{MR}^{-2}$, which is an attention guided approach specified for occluded pedestrian detection.}
\label{Table_CompFasterRCNNATT}
\end{table}

% -----------------------------------------------------------------------------------
\subsection{Performance and comparison}

%-------------------------------------------------------------------------
\begin{table}[!t]
\centering
\fontsize{7.6}{10}\selectfont
\begin{tabular}{c|c|c|c}
\hline
Method                              & \textit{Reasonable} & \textit{Heavy} & \textit{Reasonable+Heavy} \\ \hline \hline
MS-CNN \cite{DBLP:conf/eccv/CaiFFV16}& 13.22                    & 45.27               & 28.62                          \\ \hline
MultiPath \cite{DBLP:conf/bmvc/ZagoruykoLLPGCD16}& 12.19                     & 44.04               &27.42                           \\ \hline
FC-Net (Ours)                       & \textbf{11.63}               & \textbf{42.77}          & \textbf{26.21}                     \\ \hline
\end{tabular}
\caption{Comparison of context modules. For a fair comparison, all the methods use single scale features.}
\label{Table_context}
\end{table}
%-------------------------------------------------------------------------
\textbf{Citypersons dataset:} 
We compare FC-Net with state-of-the-art approaches including Adapted FasterRCNN \cite{DBLP:conf/cvpr/ZhangBS17}, Repulsion Loss \cite{wang2017repulsion}, and OR-CNN \cite{DBLP:conf/eccv/ZhangWBLL18} on the validation and test sets of CityPersons.

In Table \ref{Table_Compared_With_SOTA}, with the $\times$1.3 scale of the input image, our approach achieves 8.5$\%$ and 1.8$\%$ lower $\mathrm{MR}^{-2}$ than OR-CNN on the \textit{Heavy} subset and the \textit{Partial} subset, respectively, while maintaining a comparable performance on the \textit{Reasonable} subset. With the $\times$1 scale of the input image, it outperforms OR-CNN by 8.9$\%$ and 1.4$\%$ on the \textit{Heavy} subset and the \textit{Partial} subset, respectively.

As shown in the last column of Table \ref{Table_CityPersons_Test}, FC-Net outperforms OR-CNN up to \textbf{10.29$\%$} (41.14\% vs. 51.43\%) MR$^{-2}$ on the \textit{Heavy} subset. On the \textit{All} subset, it produces comparable performance to other approaches.

In Table\ \ref{Table_CompFasterRCNNATT}, we compare FC-Net with an attention guided approach, FasterRCNN+ATT \cite{zhang2018occluded}, which is a state-of-the-art  approach specified for occluded pedestrian detection.
Surprisingly, FC-Net outperforms FasterRCNN+ATT by \textbf{9.87\%} on the \textit{Heavy} subset and \textbf{9.59\%} on the \textit{Reasonable+Heavy} subset. It also outperforms FasterRCNN+ATT on the \textit{Reasonable} subset. We implement the attention module of FasterRCNN+ATT in the FC-Net framework (denoted as FC-Net+ATT), and find that FC-Net (ours) also outperforms FC-Net+ATT.

%-------------------------------------------------------------------------
\begin{figure}[tbp]
    \centering
    \includegraphics[width=1.0\linewidth]{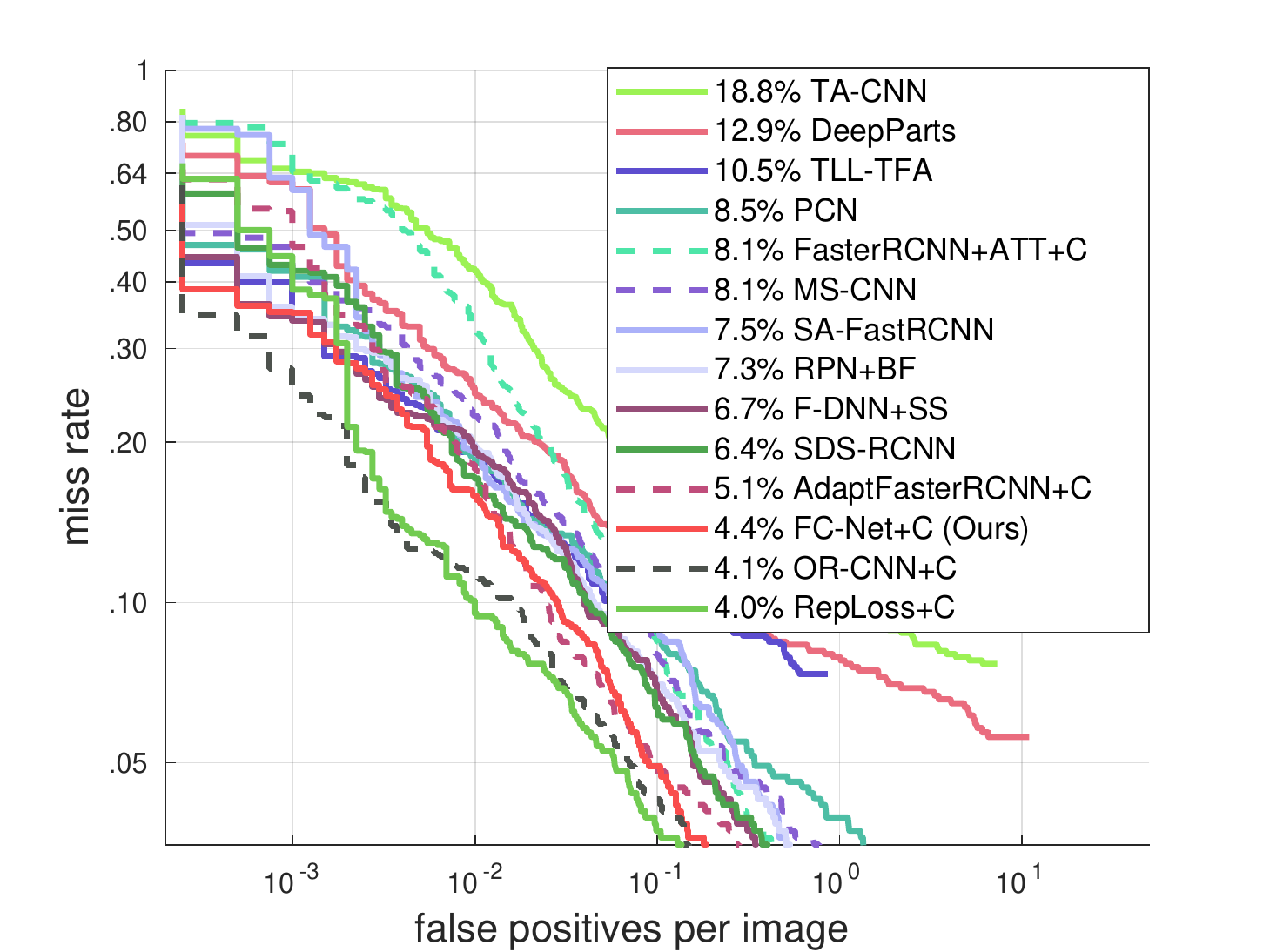}
    \caption{\label{Fig.Caltech} Comparison with state-of-the-art approaches on the Caltech dataset.
    ``C'' indicates models pre-trained on CityPersons. FC-Net achieves $4.4\%$ $\mathrm{MR}^{-2}$  and stays on the performance leading board. }
\end{figure}
%-------------------------------------------------------------------------
\begin{figure*}[t]
    \centering
    \includegraphics[width=1.0\linewidth]{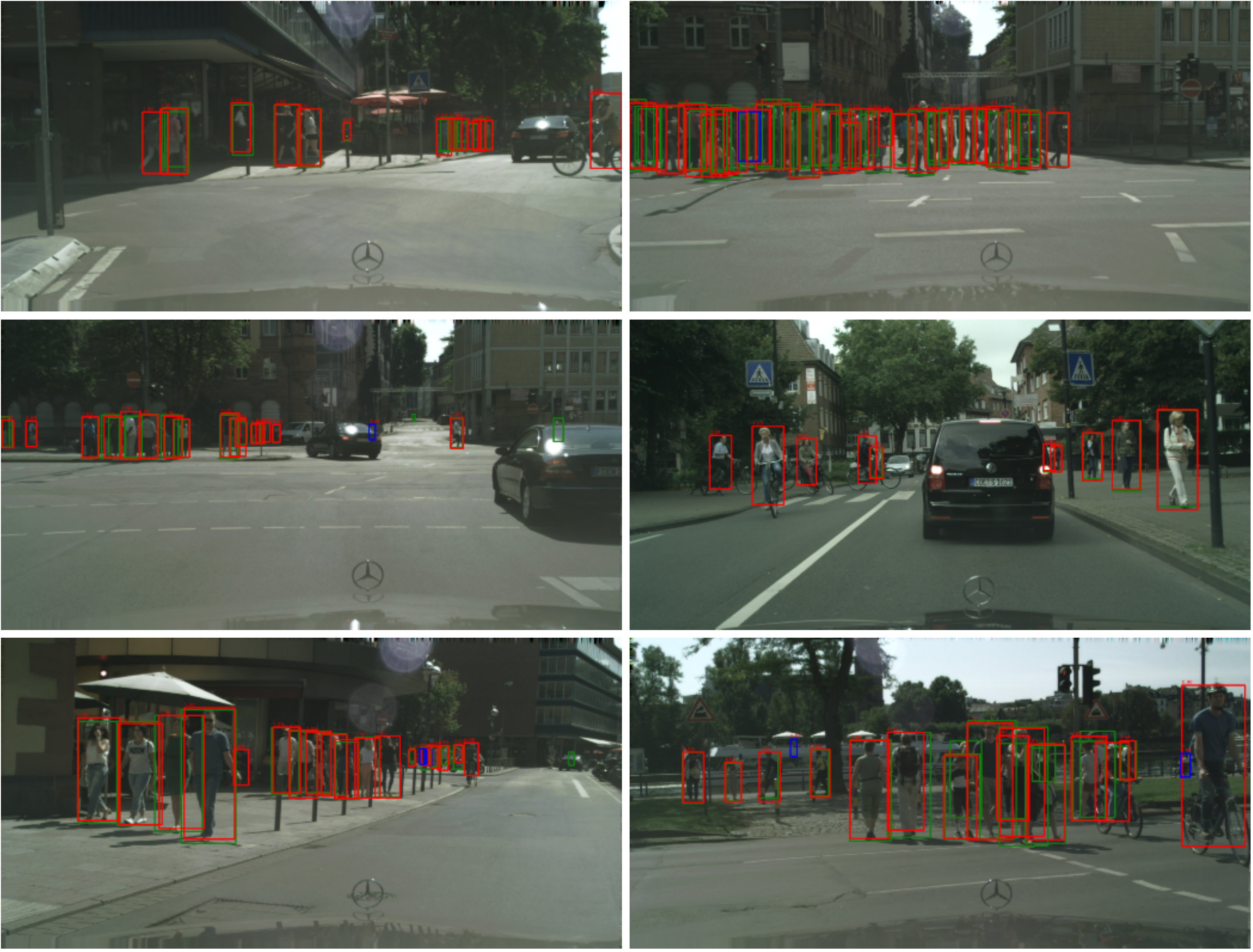}
    \caption{Examples on the CityPersons. The red, blue and green boxes indicate correctly detected pedestrians, false positives, and ground-truth, respectively.}
    \label{Fig_examples}
\end{figure*}

In Table\ \ref{Table_context}, the proposed region-calibration module is compared with the context model in MS-CNN~\cite{DBLP:conf/eccv/CaiFFV16} and MultiPath~\cite{DBLP:conf/bmvc/ZagoruykoLLPGCD16}.  It can be seen that the proposed module outperforms them. The reason lies in that our region-calibration, under the guidance of the pedestrian activation map, can adaptively produce inner and outer regions according to the pedestrian activation map. In contrast, those in MS-CNN and MultiPath are not adaptive as they use pre-defined regions.

In Fig.\ \ref{Fig_examples}, some detection examples on the CityPersons dataset are shown. We find  that FC-Net can precisely detect the pedestrians with heavy occlusions. Nevertheless, we also observe some false detections from low-resolution and/or occluded regions. The false detections could be caused by the detector's over-fitting to the few hard positives during training.
%-------------------------------------------------------------------------

\textbf{Caltech dateset:} 
On this dataset, we use the high quality annotations provided by \cite{DBLP:conf/cvpr/ZhangBOHS16}. 
Following the commonly used evaluation metric~\cite{DBLP:conf/eccv/ZhangWBLL18}, the log-average miss rate over 9 points ranging from $10^{-2}$ to $10^{0}$ FPPI is used to evaluate the performance of the detectors. 
We pre-train FC-Net on CityPersons, and then fine-tune it on the training set of Caltech.
We evaluate FC-Net on the \textit{Reasonable} subset of the Caltech dataset, and compare it to other state-of-the-art methods (\textit{e.g.}, \cite{wang2017repulsion, DBLP:conf/wacv/DuELD17, DBLP:conf/eccv/ZhangLLH16, DBLP:conf/iccv/BrazilYL17, DBLP:conf/iccv/TianLWT15, zhang2018occluded, DBLP:conf/eccv/ZhangWBLL18, DBLP:conf/eccv/CaiFFV16,  DBLP:conf/cvpr/TianLWT15}). 
As shown in Fig.\ \ref{Fig.Caltech}, FC-Net, achieving $4.4\%$ $\mathrm{MR}^{-2}$, is on the performance leading board.
%-------------------------------------------------------------------------

\subsection{General Object Detection}

In addition to pedestrian detection, the proposed FC-Net is generally applicable to other object detection. To validate it, we test FC-Net on the PASCAL VOC 2007 dataset, which consists of 20 object categories. To implement FC-Net on PASCAL VOC, we still use the Faster R-CNN framework as a baseline, and the RseNet-50 pre-trained on ImageNet as the backbone network. The model is fine-tuned on the training and validation subsets of PASCAL VOC, and is evaluated on the test subset of it. In Table \ref{Table_FC-Net_PASCAL}, FC-Net outperforms the baseline by 1.3\% mAP. Particularly, it improves the mAPs of ``aero", ``boats", ``sofa" and ``train" by 6.3\%, 5.9\%, 5.9\%, and 4.3\% respectively, which are significant improvements for the challenging object detection task. The activation maps of some categories, \textit{e.g.}, “bike”, have many “holes”, as there exist many background pixels within the object regions. This could cause a false enforcement of background features which decrease the detection performance.

\begin{figure}[t]
    \centering
    \includegraphics[width=1.0\linewidth]{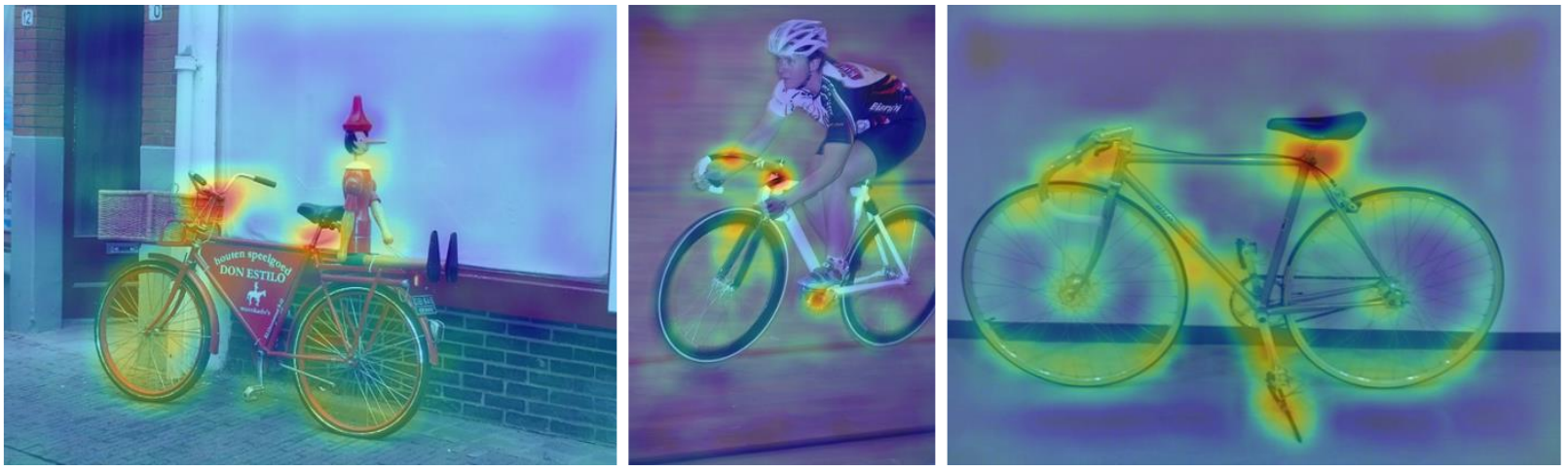}
    \caption{The activation maps of bikes.}
    \label{Fig_bikes}
\end{figure}

\subsection{Detection Efficiency} In Table \ref{Table_FC-Net_Inference_Time}, we compare the test efficiency of FC-Net with the Faster R-CNN baseline. With the superior performance on detecting occluded pedestrians, FC-Net has only a slight computational cost overhead. The SA and FC modules are called once in each training iteration. Therefore, their training iteration number is equal to that of the network. During inference, the SA and FC modules are called once with only an increment of 0.042 second per image.

%-------------------------------------------------------------------------
\begin{table*}[]
\centering
\fontsize{8.5}{10.0}\selectfont
\begin{tabular}{p{2.4cm}| @{}m{0.65cm}<{\centering}@{}m{0.65cm}<{\centering}@{}m{0.65cm}<{\centering}@{}m{0.65cm}<{\centering}@{}m{0.80cm}<{\centering}
@{}m{0.65cm}<{\centering}@{}m{0.60cm}<{\centering}@{}m{0.60cm}<{\centering}@{}m{0.65cm}<{\centering}@{}m{0.60cm}<{\centering}
@{}m{0.65cm}<{\centering}@{}m{0.65cm}<{\centering}@{}m{0.65cm}<{\centering}@{}m{0.80cm}<{\centering}@{}m{0.80cm}<{\centering}
@{}m{0.75cm}<{\centering}@{}m{0.80cm}<{\centering}@{}m{0.75cm}<{\centering}@{}m{0.65cm}<{\centering}@{}m{0.60cm}<{\centering}
@{}m{0.85cm}<{\centering}@{}}
\hline
\centering
Method                              & aero & bike & bird & boat & bottle & bus & car & cat & chair & cow & table & dog & horse & mbike & person & plant & sheep & sofa & train & tv & mAP \\ \hline \hline
Faster R-CNN \cite{DBLP:conf/nips/RenHGS15-s}        &  75.8    & \textbf{82.3}     & 72.9     & 56.0     & 61.2       & 80.6    & 85.4    & 83.7    & \textbf{55.2}      & \textbf{82.7}    & 63.2      & \textbf{82.0}    & 84.0      & 80.9      & 82.6       & \textbf{46.5}      & 73.5      & 66.8     & 75.8     & \textbf{74.2}     & 73.3      \\ \hline
FC-Net                              & \textbf{82.1}   & 79.1     & \textbf{74.9}     & \textbf{61.9}     & \textbf{63.1}       & \textbf{81.4}    & \textbf{85.8}    & \textbf{85.7}    & 55.1      & 78.9    & \textbf{65.6}      & \textbf{82.0}    & \textbf{85.6}     & \textbf{81.0}      & \textbf{82.8}       & 46.4      & \textbf{75.0}      & \textbf{72.7}    & \textbf{80.1}     & 73.4     & \textbf{74.6}      \\ \hline
\end{tabular}
\caption{\label{Table_FC-Net_PASCAL} General object detection results evaluated on PASCAL VOC 2007.}
\end{table*}

%-------------------------------------------------------------------------

\begin{table}[]
\centering
\fontsize{9}{11}\selectfont
\begin{tabular}{c|c}
\hline
               & Inference Time (second/image) \\ \hline \hline
Faster R-CNN \cite{DBLP:conf/nips/RenHGS15-s}   & 0.248              \\ \hline
FC-Net with SA & 0.283              \\ \hline
FC-Net with FC & 0.284              \\ \hline
FC-Net with SA+FC & 0.290              \\ \hline
\end{tabular}
\caption{\label{Table_FC-Net_Inference_Time} Comparison of detection speeds on CityPersons.}
\end{table}

%------------------------------------------------------------------------

\section{Conclusion}

Existing pedestrian detection approaches are unable to adapt to occluded instances while maintaining good performance on non-occluded ones. In this paper, we propose a novel feature learning method, referred to as feature calibration network (FC-Net), to adaptively detect pedestrians with heavy occlusion. FC-Net is made up of a self-activation (SA) module and a feature calibration (FC) module. The SA module estimates a pedestrian activation map by reusing the classifier weights, and the FC module calibrates the convolutional features for adaptive pedestrian representation. With the SA and FC modules, FC-Net improves the performance of occluded pedestrian detection, in striking contrast with state-of-the-art approaches. It is also applicable to general object detection tasks with significant performance gain. The underlying nature behind FC-Net is that it implements a special kind of self-paced feature learning, which can reinforce the features in visible object parts while suppressing those in occluded regions. This provides a fresh insight for pedestrian detection and other general object detection with occlusions.

\ifCLASSOPTIONcaptionsoff
  \newpage
\fi

\bibliography{egbib}
\bibliographystyle{IEEEtran}

\end{document}